\documentclass[sn-basic]{sn-jnl}

\jyear{2022}%

\theoremstyle{thmstyleone}%
\theoremstyle{thmstyletwo}%

\theoremstyle{thmstylethree}%

\makeatletter

\newcommand{\Rmnum}[1]{\expandafter\@slowromancap\romannumeral #1@}
\makeatother

\begin{document}

\title[Knowledge Graphs: Opportunities and Challenges]{Knowledge Graphs: Opportunities and Challenges}


\author[1]{\fnm{Ciyuan} \sur{Peng}}\email{ciyuan.p@outlook.com}

\author*[2]{\fnm{Feng} \sur{Xia}}\email{f.xia@ieee.org}

\author[3]{\fnm{Mehdi} \sur{Naseriparsa}}\email{m.naseriparsa@federation.edu.au}

\author[4]{\fnm{Francesco} \sur{Osborne}}\email{francesco.osborne@open.ac.uk}

\affil[1]{\orgdiv{Institute of Innovation, Science and Sustainability}, \orgname{Federation University Australia}, \orgaddress{\city{Ballarat}, \postcode{3353}, \state{VIC}, \country{Australia}}}

\affil[2]{\orgdiv{School of Computing Technologies}, \orgname{RMIT University}, \orgaddress{\city{Melbourne}, \postcode{3000}, \state{VIC}, \country{Australia}}}

\affil[3]{\orgdiv{Global Professional School}, \orgname{Federation University Australia}, \orgaddress{\city{Ballarat}, \postcode{3353}, \state{VIC}, \country{Australia}}}

\affil[4]{\orgdiv{Knowledge Media Institute}, \orgname{The Open University}, \orgaddress{\city{Milton Keynes}, \postcode{MK7 6AA}, \country{UK}}}


\abstract{With the explosive growth of artificial intelligence (AI) and big data, it has become vitally important to organize and represent the enormous volume of knowledge appropriately. As graph data, knowledge graphs accumulate and convey knowledge of the real world. It has been well-recognized that knowledge graphs effectively represent complex information; hence, they rapidly gain the attention of academia and industry in recent years. Thus to develop a deeper understanding of knowledge graphs, this paper presents a systematic overview of this field. Specifically, we focus on the opportunities and challenges of knowledge graphs. We first review the opportunities of knowledge graphs in terms of two aspects: (1) AI systems built upon knowledge graphs; (2) potential application fields of knowledge graphs. Then, we thoroughly discuss severe technical challenges in this field, such as knowledge graph embeddings, knowledge acquisition, knowledge graph completion, knowledge fusion, and knowledge reasoning. We expect that this survey will shed new light on future research and the development of knowledge graphs. }




\maketitle

\section{Introduction}\label{sec1}

Knowledge plays a vital role in human existence and development. Learning and representing human knowledge are crucial tasks in artificial intelligence (AI) research. While humans are able to understand and analyze their surroundings, AI systems require additional knowledge to obtain the same abilities and solve complex tasks in realistic scenarios \citep{ji2021survey}. To support these systems, we have seen the emergence of many approaches for representing human knowledge according to different conceptual models. In the last decade, knowledge graphs have become a standard solution in this space, as well as a research trend in academia and industry \citep{kong2022bolt}.

Knowledge graphs are defined as graphs of data that accumulate and convey knowledge of the real world. The nodes in the knowledge graphs represent the entities of interest, and the edges represent the relations between the entities~\citep{hogan2021knowledge,cheng2022financial}. These representations utilize formal semantics, which allows computers to process them efficiently and unambiguously. For example, the entity ``Bill Gates" can be linked to the entity ``Microsoft" because Bill Gates is the founder of Microsoft; thus, they have relationships in the real world.

Due to the great significance of knowledge graphs in processing heterogeneous information within a machine-readable context, a considerable amount of research has been conducted continuously on these solutions in recent years \citep{dai2020survey}. 
The proposed knowledge graphs are widely employed in various AI systems recently \citep{ko2021machine,mohamed2021biological}, such as recommender systems, question answering, and information retrieval. They are also widely applied in many fields (e.g., education and medical care) to benefit human life and society. \citep{sun2020multi,bounhas2020building}. 

Therefore, knowledge graphs have seized great opportunities by improving the quality of AI systems and being applied to various areas. 
However, the research on knowledge graphs still faces significant technical challenges. For example, there are major limitations in the current technologies for acquiring knowledge from multiple sources and integrating them into a typical knowledge graph. Thus, knowledge graphs provide great opportunities in modern society. However, there are technical challenges in their development. Consequently, it is necessary to analyze the knowledge graphs with respect to their opportunities and challenges to develop a better understanding of the knowledge graphs.

To deeply understand the development of knowledge graphs, this survey extensively analyzes knowledge graphs in terms of their opportunities and challenges. 
Firstly, we discuss the opportunities of knowledge graphs in terms of two aspects: AI systems whose performance is significantly improved by knowledge graphs and application fields that benefit from knowledge graphs.
Then, we analyze the challenges of the knowledge graph by considering the limitations of knowledge graph technologies.
The main contributions of this paper are as follows:
\begin{itemize}
	\item\textbf{Survey on knowledge graphs}. We conduct a comprehensive survey of existing knowledge graph studies. In particular, this work thoroughly analyzes the advancements in knowledge graphs in terms of state-of-the-art technologies and applications.
	\item\textbf{Knowledge graph opportunities}. We investigate potential opportunities for knowledge graphs in terms of knowledge graph-based AI systems and application fields that utilize knowledge graphs. Firstly, we examine the benefits of knowledge graphs for AI systems, including recommender systems, question-answering systems, and information retrieval. Then, we discuss the far-reaching impacts of knowledge graphs on human society by describing current and potential knowledge graph applications in various fields (e.g., education, scientific research, social media, and medical care).
	\item\textbf{Knowledge graph challenges}. We provide deep insights into significant technical challenges facing knowledge graphs. In particular, we elaborate on limitations concerning five representative knowledge graph technologies, including knowledge graph embeddings, knowledge acquisition, knowledge graph completion, knowledge fusion, and knowledge reasoning.

\end{itemize}

The rest of the paper is organized as follows. Section \ref{sec2} provides an overview of knowledge graphs, including the definitions and the categorization of existing research on knowledge graphs. To examine the opportunities of knowledge graphs, Section \ref{sec3} and Section \ref{sec4} introduce relevant AI systems and application fields, respectively. Section \ref{sec5} details the challenges of knowledge graphs based on the technologies. Finally, we conclude this paper in Section \ref{sec6}.

\section{Overview}\label{sec2}

In this section, the definition of knowledge graphs is provided first; then, we categorize significant state-of-the-art research in this area.

\subsection{What are Knowledge Graphs?}

\begin{figure}[!t]
	
	\centering
	\includegraphics[scale=0.5]{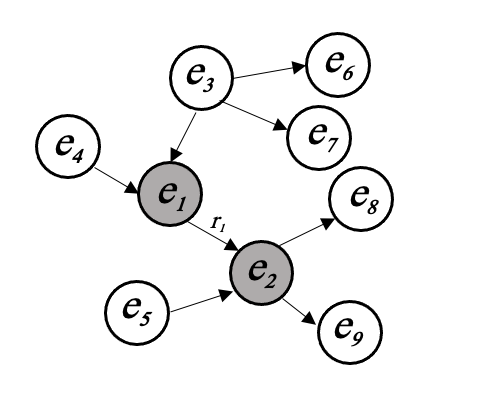}
	\caption{An example of a knowledge graph. In this knowledge graph, $(e_1,r_1,e_2)$ is a triplet that indicates $e_1$ and $e_2$ are connected by relation $r_1$.}
	\label{1}
\end{figure}

A knowledge base is a typical data set that represents real-world facts and semantic relations in the form of triplets. When the triplets are represented as a graph with edges as relations and nodes as entities, it is considered a knowledge graph. Generally, the knowledge graph and knowledge base are regarded as the same concept and are used interchangeably. In addition, the schema for a knowledge graph can be defined as an ontology, which shows the properties of a specific domain and how they are related. Therefore, one essential stage of knowledge graph construction is ontology construction.

In 2012, Google first put forward Knowledge Graph by introducing their knowledge base called \textit{Google Knowledge Graph} \citep{ehrlinger2016towards}. Afterward, many knowledge graphs are introduced and adopted such as:

\begin{itemize}
	\item \textit{DBpedia}, a knowledge graph that intends to discover semantically meaningful information form Wikipedia and convert it into an effective well-structured ontological knowledge base in DBpedia \citep{auer2007dbpedia}.
	
	\item \textit{Freebase}, a knowledge graph which is built upon multiple sources that provides a structured and global resource of information \citep{bollacker2008freebase}.
	
	\item \textit{Facebook’s entity graph}, a knowledge graph that converts the unstructured content of the user profiles into meaningful structured data \citep{ugander2011anatomy}.
	
	\item \textit{Wikidata}, a cross-lingual document-oriented knowledge graph which supports many sites and services such as Wikipedia \citep{vrandevcic2014wikidata}.
	
	\item \textit{Yago}, is a quality knowledge base that contains a huge number of entities and their corresponding relationships. These entities are extracted from multiple sources such as Wikipedia and WordNet \citep{rebele2016yago}.
	
	\item \textit{WordNet}, is a lexical knowledge base to measure the semantic similarity between words. The knowledge base contains a number of hierarchical concept graphs to analyse the semantic similarity \citep{pedersen2004wordnet}.   
\end{itemize}

A knowledge graph is a directed graph composed of nodes and edges, where one node indicates an entity (a real object or abstract concept), and the edge between the two nodes conveys the semantic relation between the two entities \citep{bordes2011learning}. Resource Description Framework (RDF) and Labeled Property Graphs (LPGs) are two typical ways to represent and manage knowledge graphs \citep{farber2018linked,baken2020linked}.
The fundamental unit of a knowledge graph is the triple \emph{(subject, predicate, object)} (or \emph{(head, relation, tail)}), i.e., \emph{(Bill Gates, founderOf, Microsoft)}. Since the relation is not necessarily symmetric, the direction of a link matters.
Therefore, a knowledge graph can also be seen as a directed graph in which the head entities point to the tail entities via the relation's edge. 

Fig.~\ref{1} depicts an example of a simple knowledge graph. As shown in Fig.~\ref{1}, nodes $e_1$ and $e_2$ darkened in color are connected by relation $r_1$, which goes from $e_1$ to $e_2$. Therefore, $e_1$, $e_2$, and $r_1$ can form the triplet $(e_1,r_1,e_2)$, in which $e_1$ and $e_2$ are the head and tail entities, respectively.

\subsection{Current Research on Knowledge Graphs }

\begin{figure*}[!t]
	
	\centering
	\includegraphics[scale=0.45]{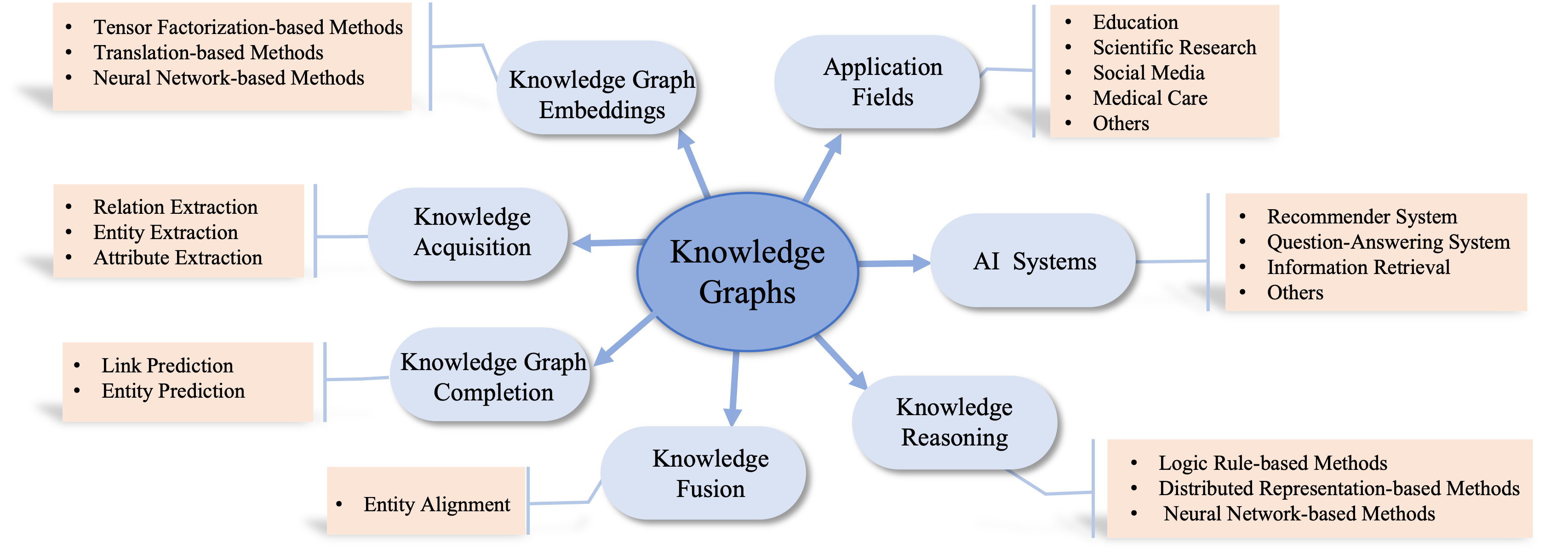}
	\caption{Research on knowledge graphs.}
	\label{22}
\end{figure*}

In recent years, knowledge graphs have gained extensive research interest. Plenty of studies have focused on exploring knowledge graphs. This paper conducts a comprehensive survey on knowledge graphs and lists seven important categories of current research on this topic. Fig.~\ref{22} illustrates a schema of the most popular research lines regarding knowledge graphs. Among them, AI systems are services that utilize knowledge graphs for their foundation, and application fields are domains where knowledge graphs reach. These two research lines are listed for discussing the opportunities of knowledge graphs. Another five research lines are five main knowledge graph technologies corresponding to five tasks. In this paper, we introduce these five technologies and emphasize their limitations to give useful insights into the major challenges of the knowledge graphs.

\textbf{Knowledge Graph Embedding:} Knowledge graph embedding is one of the central research issues. This task aims to map entities and relations of a knowledge graph to a low-dimensional vector space so that it captures the semantics and the structure of the knowledge graph efficiently \citep{dai2020survey}. Then, the obtained feature vector can be effectively learned by machine learning models. 
Three main triplet fact-based embedding methods are as follows: (a) tensor factorization-based, (b) translation-based, and (c) neural network-based methods \citep{dai2020survey}.


\textbf{Knowledge Acquisition:} Knowledge acquisition, which focuses on modeling and constructing knowledge graphs, is another crucial research direction of knowledge graph study.
Typically, the knowledge is imported from structured sources by employing mapping languages, such as R2RML \citep{rodriguez2015efficient}. Furthermore, the knowledge could be extracted from unstructured documents (e.g., news, research papers, and patents) by adopting relation, entity, or attribute extraction methods \citep{Liu2020SIGIR,yu2020relationship,yao2019kg}.

\textbf{Knowledge Graph Completion:} Although there are many methods for constructing knowledge graphs, it is still unfeasible to create comprehensive representations of all the knowledge in a field. 
Most knowledge graphs still lack a good number of entities and relationships. Thereby, significant efforts have been made for completing knowledge graphs. Knowledge graph completion aims to improve the quality of knowledge graphs by predicting additional relationships and entities. The first task typically adopts link prediction techniques to generate triplets and then assigns the triplets plausibility scores \citep{ji2021survey}. 
The second task employs entity prediction methods for obtaining and integrating further information from external sources. 

\textbf{Knowledge Fusion:} Knowledge fusion is also an important research direction that focuses on capturing knowledge from different sources and integrating it into a knowledge graph \citep{nguyen2020knowledge}. The knowledge fusion approaches are useful for both generating and completing knowledge graphs. Recently, entity alignment has been the primary method for implementing knowledge fusion tasks.

\textbf{Knowledge Reasoning:} Tremendous research efforts have focused on reasoning to enrich the knowledge graphs, which aims to infer new facts based on existing data \citep{minervini2020differentiable}. In particular, new relations between two unconnected entities are inferred, forming new triplets. Also, by reasoning out the false facts, knowledge reasoning has the ability to identify erroneous knowledge. The main methods for knowledge reasoning include logic rule-based, distributed representation-based, and neural network-based methods \citep{chen2020review}.

\textbf{AI Systems:}
Nowadays, knowledge graphs are widely utilized by AI systems \citep{liang2022aspect}, such as recommenders, question-answering systems, and information retrieval tools. Typically, the richness of information within knowledge graphs enhances the performance of these solutions. Therefore, many studies have focused on taking advantage of knowledge graphs to improve AI systems' performance.

\textbf{Application Fields:} Knowledge graphs have numerous applications in various fields, including education, scientific research, social media, and medical care \citep{li2020real}. A variety of intelligent applications are required to improve the standard of human life.

Differing from other works, this paper focuses on surveying the opportunities and challenges of knowledge graphs. 
In particular, knowledge graphs meet great opportunities by improving the quality of AI services and being applied in various fields. On the contrary, this paper regards the limitations of knowledge graph technologies as the challenges. Therefore, we will discuss the technical limitations regarding knowledge graph embeddings, knowledge acquisition, knowledge graph completion, knowledge fusion, and knowledge reasoning.

\section{Knowledge Graphs for AI Systems}\label{sec3}

This section explains the opportunities by analyzing the advantages that knowledge graphs bring for improving the functionalities of AI Systems. Specifically, there are a couple of systems, including recommender systems, question-answering systems, and information retrieval tools \citep{2020dA,zou2020}, which utilize knowledge graphs for their input data and benefit the most from knowledge graphs. In addition to these systems, other AI systems, such as image recognition systems \citep{chen2020knowledge}, have started to consider the characteristic of knowledge graphs. However, the application of knowledge graphs in these systems is not widespread. Moreover, these systems do not directly optimize performance by utilizing knowledge graphs for the input data. Therefore, the advantages that knowledge graphs bring for recommender systems, question-answering systems, and information retrieval tools are discussed in detail to analyze the opportunities of knowledge graphs. Typically, these solutions greatly benefit from adopting knowledge graphs that offer high-quality representations of the domain knowledge. Table~\ref*{t_services} presents a summary of the AI systems that we will discuss below.

\begin{sidewaystable}
	\sidewaystablefn%
	\begin{center}
		\begin{minipage}{\textheight}
			
	\caption{AI systems using knowledge graphs.}
	\label{t_services}

	\begin{tabular}{p{0.25\columnwidth}p{0.25\columnwidth}p{0.4\columnwidth}}
		\toprule
		AI Systems & Approaches &Techniques on knowledge graphs\\
		\midrule
	Recommender systems& KPRN \citep{wang2019explainable}& Entity-relation path generation based on user-item interaction\\ 
		& RippleNet \citep{wang2018ripplenet}& Preference propagation\\
		 & MKR \citep{wang2019multi}& Laten user-item interaction\\
		& MKGAT \citep{sun2020multi}&Neighbor information extraction; relation reasoning\\
		 & Ripp-MKR \citep{wang2021multitask}& Preference propagation; laten user-item interaction\\
		 & RKG \citep{shu2021user}& User preferenfce lists-based knowledge graph construction\\
		Question-answering systems&  MHPGM \citep{bauer2018commonsense}& Multiple hop relation reasoning \\
		& PCQA \citep{shin2019predicate} &Predicate constraints-based relation extraction\\
		& KEQA \citep{huang2019knowledge}&Simple question-based triplet construction \\
		& EmbedKGQA \citep{saxena2020improving}& Knowledge graph embedding-based multi-hop question answering\\
		Information retrieval& EQFE \citep{dalton2014entity}&Query knowledge graph-based feature expansion\\
		& Knowledge graph based Information Retrieval Technology \citep{wang2018information}& Query-document knowledge graph construction\\
		& CKG  \citep{wise2020covid}&Document knowledge graph construction\\
		& EDRM\citep{liu2018entity}&Integration of semantics from knowledge graphs and entities from queries and documents
		representations of their entities\\

		\bottomrule

	\end{tabular}
		\end{minipage}
\end{center}
\end{sidewaystable}

\subsection{Recommender Systems}\label{subsec2}

With the continuous development of big data, we observe the exponential growth of information. In the age of information explosion, it becomes challenging for people to receive valid and reliable information \citep{shokeen2020study,monti2021systematic,gomez2022enabling}. Specifically, online users may feel confused when they want to select some items they are interested in among thousands of choices. To tackle this issue, we saw the emergence of several recommender systems to provide users with more accurate information. 
Typically, recommender systems learn the preference of target users for a set of items \citep{Wan2020TNSE,zheng2022survey} and produce a set of suggested items with similar characteristics. 
Recommender systems are fruitful solutions to the information explosion problem and are employed in various fields for enhancing user experience \citep{2020Recommender}.

\subsubsection{Traditional Recommender Systems}\label{subsubsec2}

There are two traditional methods for developing recommender systems, including content-based and collaborative filtering-based (CF-based) methods. Shu et al. \citep{sun2019research} and Guo et al. \citep{2020dA} have compared and summarised these two approaches. 

\textbf{Content-based Recommender Systems}: The content-based recommender systems first analyze the content features of items (e.g., descriptions, documents). These items are previously scored by the target users \citep{2020dA,Xia2014WWWmulti}. Then, the recommender systems learn the user interests by employing machine learning models. Thus, these systems are able to effectively recommend trending items to the target users according to their preferences. Some recommender systems utilize the content of the original query result to discover highly-related items for the users that may interest them \citep{NaseriparsaLIZ19}. These systems employ machine learning techniques or statistical measures such as correlation to compute the highly-similar items to those that are visited by the users \citep{NaseriparsaILC19}. Another group of content-based recommender systems employs lexical references such as dictionaries to utilize semantic relationships of the user query results to recommend highly semantically-related items to the users that may directly satisfy their information needs \citep{NaseriparsaILM18, SunX0L17}. 

\textbf{CF-based Recommender Systems}: CF-based recommender systems suggest items to the users based on the information of user-item interaction \citep{2020A}. CF-based recommender systems infer the user preference by clustering similar users instead of extracting the features of the items \citep{Wang2019TCSSsustainable}. However, we face data sparsity and cold start problems in traditional CF-based systems. In general, users can only rate a few items among a large number of items, which leads to preventing many items from receiving appropriate feedback. Therefore, the recommender systems do not effectively learn user preferences accurately because of data sparsity \citep{Bai2019IACCESSscientific, Xia2014ISJsocially}. On the other hand, the cold start problem makes it even more difficult to make recommendations when the items or users are new because there is no historical data or ground truth.
Moreover, because abundant user information is required for achieving effective recommendations, CF-based recommender systems face privacy issues. How to achieve personalized recommendations while protecting the privacy of users is still an unsolved problem.

\subsubsection{Knowledge Graph-based Recommender Systems}

 To address inherent problems of traditional approaches, the community has produced several hybrid recommender systems, which consider both item features and the distribution of user scores. Most of these solutions adopt knowledge graphs for representing and interlinking items \citep{palumbo2020entity2rec}. Specifically, Knowledge graph-based recommender systems integrate knowledge graphs as auxiliary information and leverage users and items networks to learn the relationships of items-users, items-items, and users-users \citep{palumbo2018knowledge}.

\begin{figure}[!t]
	
	\centering
	\includegraphics[scale=0.45]{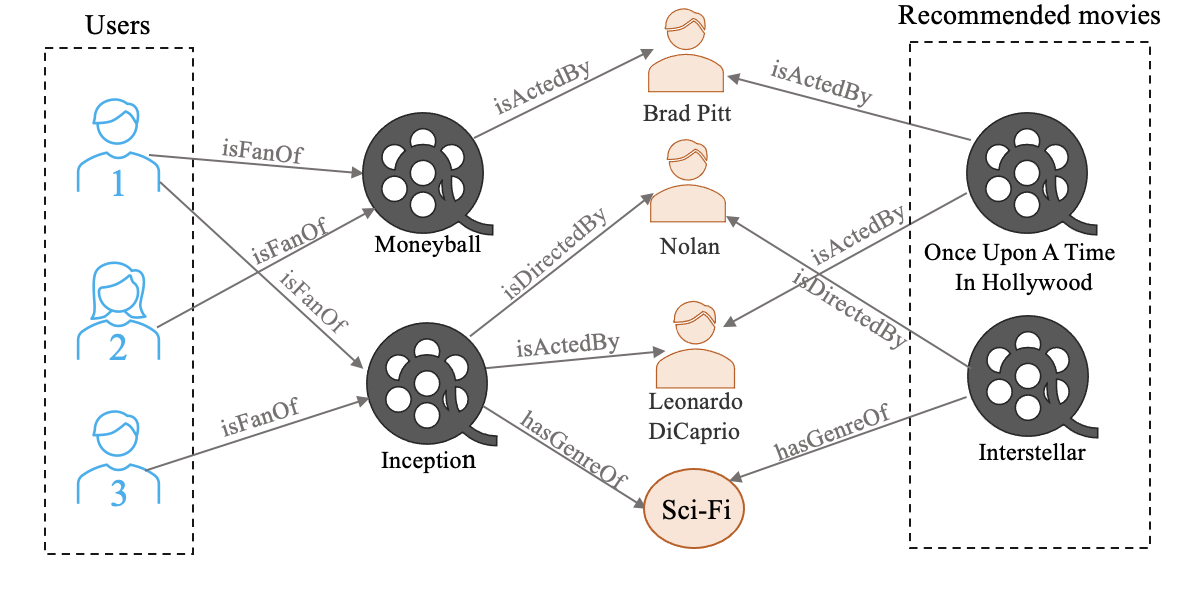}
	\caption{An example of knowledge graph-based recommender system.}
	\label{3}
\end{figure}

Fig~\ref{3} presents an example of knowledge graph-based movie recommendation. Here we can see that the movies ``Once Upon A Time in Hollywood" and ``Interstellar" are recommended to three users according to a knowledge graph that contains the nodes of users, films, directors, actors, and genres. The knowledge graph is thus used to infer latent relations between the user and the recommended movies. 

Recently, a great deal of research has been conducted to utilize knowledge graphs for recommendation tasks. For instance, Wang et al. \citep{wang2019explainable} introduced KPRN. KPRN is a recommender system that generates entity-relation paths according to the user-item interaction and constructs a knowledge graph that consists of the users, items, and their interaction. It then infers the user preference based on the entity-relation path. The user-item interaction, which is extracted from knowledge graphs, improves the quality of the recommendations and allows the presentation of the recommended results in a more explainable manner. Wang et al. \citep{wang2019multi} also applied multi-task knowledge graph representation (MKR) for recommendation tasks. MKR models knowledge graphs based on the user-item interaction. It is worth noting that MKR focuses on the structural information of knowledge graphs for learning the latent user-item interaction. Sun et al. \citep{sun2020multi} proposed a Multi-modal Knowledge Graph Attention Network (MKGAT) for achieving precise recommendations. MKGAT constructs knowledge graphs based on two aspects: (1) it enriches entity information by extracting the information of the neighbor entities; (2) it scores the triplets to construct the reasoning relations. Finally, they applied knowledge graphs that are enriched with structured data to recommender systems. 

Wang et al. \citep{wang2018ripplenet} presented the RippleNet model, which incorporates knowledge graphs into recommendation tasks by preference propagation. RippleNet firstly regards users' historical records as the basis of a knowledge graph. Then, it predicts the user preference list among candidate items based on the knowledge graph links. Based on both RippleNet and MKR models, Wang et al. \citep{wang2021multitask} applied the Ripp-MKR model. Ripp-MKR combines the advantages of preference propagation and user-item interaction to dig the potential information of knowledge graphs. 
Shu et al. \citep{shu2021user} proposed RKG, which achieves recommendation by referring to the user preference-based knowledge graph. RKG first obtains users' preference lists; then, it analyzes the relations between the user's preferred items and the items which are to be recommended. Therefore, the model effectively learns the score of the candidate items for recommendation according to the candidate items' relationship with the user's preferred items.

Many studies have utilized ontological knowledge base information to improve retrieving results from various data sources \citep{FarfanHRW09}. Wu et al. \citep{WuYY13} adopted the ontological knowledge base to extract highly semantically similar sub-graphs in graph databases. Their method effectively recommends semantically relevant sub-graphs according to ontological information. Farf et al. \citep{FarfanHRW09} proposed the XOntoRank, which adopts the ontological knowledge base to facilitate the data exploration and recommendation on XML medical records.    

Compared with the traditional recommender systems, knowledge graph-based recommender systems have the following advantages:
\begin{itemize}
	\item\textbf{Better Representation of Data:} Generally, the traditional recommender systems suffer from data sparsity issues because users usually have experience with only a small number of items. However, the rich representation of entities and their connections in knowledge graphs alleviate this issue. 
	
	\item\textbf{Alleviating Cold Start Issues:} It becomes challenging for traditional recommender systems to make recommendations when there are new users or items in the data set. In knowledge graph-based recommender systems, information about new items and users can be obtained through the relations between entities within knowledge graphs. For example, when a new Science-Fiction movie such as ``Tenet'' is added to the data set of a movie recommender system that employs knowledge graphs, the information about ``Tenet" can be gained by its relationship with the genre Science-Fiction (gaining triplet \emph{(Tenet, has genre of, Sci-Fi)}). 
	
	\item\textbf{The Explainability of Recommendation:} Users and the recommended items are connected along with the links in knowledge graphs. Thereby, the reasoning process can be easily illustrated by the propagation of knowledge graphs.
\end{itemize}

\subsection{Question-answering Systems}
Question answering is one of the most central AI services, which aims to search for the answers to natural language questions by analyzing the semantic meanings \citep{dimitrakis2020survey,das2022improvement}. 
The traditional question-answering systems match the textual questions with the answers in the unstructured text database. In the search process, the semantic relationship between the question and answer is analyzed; then, the system matches the questions and answers with the maximum semantic similarity. Finally, the system outputs the answer. However, the answers are obtained by filtrating massive unstructured data, which deteriorates the efficiency of the traditional question-answering systems due to analyzing an enormous search space. To solve this issue, a lot of research focuses on employing structured data for question answering, particularly knowledge graph-based question-answering systems \citep{singh2020no,qiu2020stepwise}.

\begin{figure}[!t]
	
	\centering
	\includegraphics[scale=0.4]{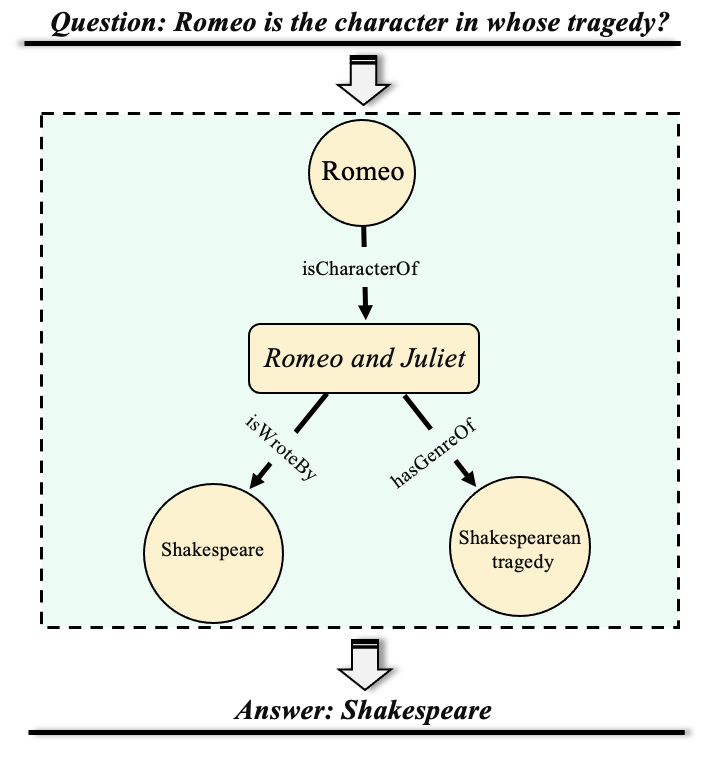}
	\caption{The illustration of knowledge graph based question-anwsering systems.}
	\label{4}
\end{figure}

The sophisticated representation of information in knowledge graphs is a natural fit for question-answering systems. Knowledge graph-based question-answering systems typically analyze the user question and retrieve the portion of knowledge graphs for answering. The answering task is facilitated either by using similarity measures or by producing structured queries in standard formats (e.g., SPARQL). 
Fig~\ref{4} presents an example of the knowledge graph-based question-answering system. The system answer ``Shakespeare" is a node that is linked to the node ``Romeo". The node ``Romeo" is extracted from the question. 

There are two main types of questions in this space: simple and multi-hop questions, respectively. Simple questions are answered only by referring to a single triplet, while multi-hop questions require combining multiple entities and relations. Focusing on simple questions, Huang et al. \citep{huang2019knowledge} proposed a knowledge graph embedding-based question-answering system (KEQA). They translated the question and its corresponding answer into a single triplet. For instance, the question `` Which film acted by Leonardo" and one of its answers ``Inception" can be expressed as the following triplet: $(Leonard, act, Inception)$. Then, the head entity, relation, and tail entity of the triplet are represented by a vector matrix in the embedding space for learning the question-answer information. Considering the semantic meanings of the questions, Shin et al. \citep{shin2019predicate} presented a predicate constraint-based question-answering system (PCQA). They took advantage of the predicate constraints of knowledge graphs, which is a triplet contains a subject, predicate, and an object to capture the connection between the questions and answers. Using the triplet for question-answering integration, the processing of the question-answering service can be simplified; therefore, the result improves.

Bauer et al. \citep{bauer2018commonsense} focused on multi-hop questions and proposed a Multi-Hop Pointer-Generator Model (MHPGM). They selected the relation edges that are related to the questions in a knowledge graph and injected attention to achieve multi-hop question answering. Because of the advantages of knowledge graphs' structure, multi-hop question answering can extract coherent answers effectively.
Saxena et al. \citep{saxena2020improving} proposed EmbedKGQA to achieve multi-hop question answering over sparse knowledge graphs (such as knowledge graphs with missing edges). The main idea of EmbedKGQ is to utilize knowledge graph embeddings to reduce knowledge graph sparsity. It first creates embeddings of all entities and then selects the embedding of a given question. Lastly, it predicts the answer by combining these embeddings.

Compared to the traditional question answering, the advantages of knowledge graph-based question-answering systems can be summarized as follows:

\begin{itemize}
	\item\textbf{Increased Efficiency:} Instead of searching for answers from massive textual data, which may contain a large volume of useless data items, knowledge graph-based question-answering systems focus only on entities with relevant properties and semantics. Therefore, they reduce the search space significantly and extract the answers effectively and efficiently. 
	
	\item\textbf{Multi-hop  Question Answering:} The answers can be more complex and sophisticated than the ones produced with traditional methods relying on unstructured data since they can combine several facts and concepts from the knowledge graph via multi-hop question answering.
\end{itemize}

\subsection{Information Retrieval}

Information retrieval enables retrieval systems to match end-user queries with relevant documents, such as web pages \citep{Liu2019CSRdata}. Traditional information retrieval systems index the documents according to the user queries and return the matched documents to the users \citep{hersh2021information}. Nevertheless, index processing is complex and requires plenty of time because of the massiveness and diversity of documents. As a result, traditional information retrieval faces the challenge of inaccurate search results and potentially low efficiency. Also, since search engines have limitations with respect to text interpretation ability, keyword-based text search usually outputs limited results. Thus, to address these problems, many modern search engines take advantage of knowledge graphs \citep{bounhas2020building,zheng2020dgl}.
Knowledge graph-based information retrieval introduces a new research direction that takes advantage of knowledge graphs for improving the performance of search engines and the explainability of the results.

Typically, these systems rely on the advanced representation of the documents based on entities and relationships from knowledge graphs. These formal and machine-readable representations are then matched to the user query for retrieving the more pertinent documents. 
For instance, Wise et al. \citep{wise2020covid} proposed a COVID-19 Knowledge Graph (CKG) to extract the relationships between the scientific articles about COVID-19. In particular, they combined the topological information of documents with the semantic meaning to construct document knowledge graphs. 
Wang et al. \citep{wang2018information} proposed a knowledge graph-based information retrieval technology that extracts entities by mining entity information on web pages via an open-source relation extraction method. Then, the entities with relationships are linked to construct a knowledge graph.

Knowledge graphs can also support methods for query expansion, which is able to enrich the user query by adding relevant concepts (e.g., synonymous).
For example, Dalton et al. \citep{dalton2014entity} presented an entity query feature
expansion (EQFE) to enrich the queries based on the query knowledge graph, including structured attributes and text. 
Liu et al. \citep{liu2018entity} proposed the Entity-Duet Neural Ranking Model (EDRM).
EDRM integrates the semantics extracted from knowledge graphs with the distributed representations of entities in queries and documents. Then, it ranks the search results using interaction-based neural ranking networks.

Compared to traditional information retrieval, the knowledge graph-based information retrieval has the following advantages:
\begin{itemize}
	\item\textbf{Semantic Representation of Items:} Items are represented according to a formal and interlinked model that supports semantic similarity, reasoning, and query expansion. This typically allows the system to retrieve more relevant items and makes the system more interpretable.
	\item\textbf{High Search Efficiency:} Knowledge graph-based information retrieval can use the advanced representation of the items to reduce the search space significantly (e.g., discarding documents that use the same terms with different meanings), resulting in improved efficiency. 

	\item\textbf{Accurate Retrieval Results:} In knowledge graph-based information retrieval, the correlation between query and documents is analyzed based on the relations between entities in the knowledge graph. This is more accurate than finding the similarities between queries and documents.

\end{itemize}

\section{Applications and Potentials}\label{sec4}

In this section, we discuss the applications and potentials of knowledge graphs in four domains: education, scientific research, social networks, and health/medical care. Although some researchers try to take advantage of knowledge graphs to develop beneficial applications in other domains such as finance \citep{CHENG2022108218}, the knowledge graph-based intelligent service in these areas is relatively obscure and still needs to be explored. Therefore, this section mainly focuses on education, scientific research, social networks, and medical care to summarize the opportunities of knowledge graphs. Table ~\ref*{t_application} presents several recent applications of knowledge graphs that make contributions to these fields.

\begin{sidewaystable}
	\sidewaystablefn%
	\begin{center}
		\begin{minipage}{\textheight}
			\caption{Fields of applications of knowledge graphs.}\label{t_application}

	\begin{tabular}{p{0.15\columnwidth}p{0.2\columnwidth}p{0.25\columnwidth}p{0.3\columnwidth}}
		\toprule
		Fields&Applications &Methods&Functions\\
		\midrule
		Education&Knowledge Graph based Course Management Model \citep{aliyu2020development} &Course knowledge graphs& Courses management; Generation of course allocation schedule\\ 
		& KnowEdu \citep{chen2018knowedu}&Instructional concepts extraction; Educational relation identification&Educational knowledge graph construction\\
		& Knowledge Graph-based Tool for Online Learning  \citep{zablith2022constructing}&Integration of social media contents and formal learning contents &Efficient online knowledge acquisition\\
		Scientific Research&Scientific Publication Management Model \citep{chi2018knowledge}& Knowledge graph based academic network &Scientific publication management\\
		&Reviewer Recommendation System \cite{9404099}& Knowledge graph-based rule engine establishment& Precise matching of reviewer and paper\\
		Social Networks&  DEAP-FAKED \citep{mayank2021deap}&News-Entity knowledge graphs& Fake news detection\\
		& GraphRec \citep{fan2019graph} & Information aggregation of user-user and user-item graphs&Social Recommendation\\
		& Graph Reasoning Model \citep{wang2018deep} & Knowledge graph propogation&Social relationship extraction\\
		Health/Medical Care& SMR \citep{gong2021smr}&Medical knowledge graph embeddings&Safe medicine recommendation\\
		& DETERRENT \citep{cui2020deterrent}&Knowledge guided graph attention network&Health misinformation detection\\
		& KGNN \citep{lin2020kgnn}&Mining the relationships between drugs&Drug discovery\\
		& COVID-KG\citep{yuan2021doctor}&Multimedia knowledge graph construction&Drug discovery\\
		
		\bottomrule
		
	\end{tabular}
		\end{minipage}
\end{center}
\end{sidewaystable}

\subsection{Education}

Education is of great importance to the development of human society. Many studies have focused on deploying intelligent applications to improve the quality of education \citep{Bai2021ebd,Wang2020TWEB}. 
Specifically, in the age of big data, data processing becomes a challenging task because of the complex and unstructured educational data. Thereby, intelligent educational systems tend to apply structured data, such as knowledge graphs. Several knowledge graph-based applications support the educational process, focusing in particular on data processing and knowledge dissemination \citep{yao2020joint}. 

In education, the quality of offline school teaching is of vital importance. Therefore, several knowledge graph-based applications focus on supporting teaching and learning. For example, considering the importance of course allocation tasks in university, Aliyu et al. \citep{aliyu2020development} proposed a knowledge graph-based course management approach to achieve automatic course allocation. They constructed a course knowledge graph in which the entities are courses, lecturers, course books, and authors in order to suggest relevant courses to students. 
Chen et al.\citep{chen2018knowedu} presented KnowEdu, a system for educational knowledge graph construction, which automatically builds knowledge graphs for learning and teaching in schools. First, KnowEdu extracts the instructional concepts of the subjects and courses as the entity features. Then, it identifies the educational relations based on the students' assessments and activities to make the teaching effect more remarkable. 

The abovementioned knowledge graph-based intelligent applications are dedicated to improving the quality of offline school teaching. However, online learning has become a hot trend recently. Moreover, online study is an indispensable way of learning for students during the COVID-19 pandemic\citep{saraji2022extended}. Struggling with confusing online content (e.g., learning content of low quality on social media), students face major challenges in acquiring significant knowledge efficiently. Therefore, researchers have focused on improving online learning environments by constructing education-efficient knowledge graphs \citep{d2016use,pereira2017linked}. For example, to facilitate online learning and establish connections between formal learning and social media, Zablith \citep{zablith2022constructing} proposed to construct a knowledge graph by integrating social media and formal educational content, respectively. Then, the produced knowledge graph can filter social media content, which is fruitful for formal learning and help students with efficient online learning to some extent.

Offline school teaching and online learning are two essential parts of education, and it is necessary to improve the quality of both to promote the development of education. Significantly, knowledge graph-based intelligent applications can deal with complicated educational data and make both offline and online education more convenient and efficient.

\subsection{Scientific Research}

A variety of knowledge graphs focus on supporting the scientific process and assisting researchers in exploring research knowledge and identifying relevant materials \citep{Xia2016TBDscientific}.
They typically describe documents (e.g., research articles, patents), actors (e.g., authors, organizations), entities (e.g., topics, tasks, technologies), and other contextual information (e.g., projects, funding) in an interlinked manner. For instance, Microsoft Academic Graph (MAG)~\citep{wang2020microsoft}
is a heterogeneous knowledge graph. MAG contains the metadata of more than 248M scientific publications, including citations, authors, institutions, journals, conferences, and fields of study. The AMiner Graph~\citep{zhang2018name} is the corpus of more than 200M publications generated and used by the AMiner system\footnote{AMiner - \url{https://www.aminer.cn/}}. The Open Academic Graph (OAG)\footnote{Open Academic Graph - \url{https://www.openacademic.ai/oag/}} is a massive knowledge graph that integrates Microsoft Academic Graph and AMiner Graph.
AceKG~\citep{10.1145/3269206.3269252} is a large-scale knowledge graph that provides 3 billion triples of academic facts about papers, authors, fields of study, venues, and institutes, as well as the relations among them. 
The Artificial Intelligence Knowledge Graph (AI-KG)~\citep{dessi2020AIKG}\footnote{AI-KG - \url{https://w3id.org/aikg/}} describes 800K entities (e.g., tasks, methods, materials, metrics) extracted from the 330K most cited articles in the field of AI. 
The Academia/Industry Dynamics Knowledge Graph (AIDA KG)~\citep{angioni2021aida}\footnote{AIDA - \url{http://w3id.org/aida}} describes 21M publications and 8M patents according to the research topics drawn from the Computer Science Ontology~\citep{salatino2020CSO} and 66 industrial sectors (e.g., automotive, financial, energy, electronics).

In addition to constructing academic knowledge graphs, many researchers also take advantage of knowledge graphs to develop various applications beneficial to scientific research.
Chi et al. \citep{chi2018knowledge} proposed a scientific publication management model to help non-researchers learn methods for sustainability from research thinking. They built a knowledge graph-based academic network to manage scientific entities. The scientific entities, including researchers, papers, journals, and organizations, are connected regarding their properties. For the convenience of researchers, many scientific knowledge graph-based recommender systems, including citation recommendation, collaboration recommendation, and reviewer recommendation, are put forward \citep{shao2021survey}. For instance, Yong et al.\citep{9404099} designed a knowledge graph-based reviewer assignment system to achieve precise matching of reviewers and papers. Particularly, they matched knowledge graphs and recommendation rules to establish a rule engine for the recommendation process.

\subsection{Social Networks}

With the rapid growth of social media such as Facebook and Twitter, online social networks have penetrated human life and bring plenty of benefits such as social relationship establishment and convenient information acquisition \citep{Li2020ISYS,hashemi2020multi}. Various social knowledge graphs are modeled and applied to analyze the critical information from the social network. These knowledge graphs are usually constituted based on the people's activities and their posts on social media, which are applied to numerous applications for different functions \citep{Xu2020IJCDL}.

Remarkably, social media provides high chances for people to make friends and gain personalized information. Furthermore, social media raises fundamental problems, such as how to recommend accurate content that interests us and how to connect with persons interested in a common topic. To address these issues, various studies have been proposed to match users with their favorite content (or friends) for recommendation \citep{ying2018graph}. With the increase in users' demand, a number of researchers utilize knowledge graph-based approaches for more precise recommendations \citep{gao2020deep}. 
A representative example is GraphRec (a graph neural network framework for social
recommendations) proposed by Fan et al. \citep{fan2019graph}. They considered two kinds of social knowledge graphs: user-user and user-item graphs. Then, they extracted information from the two knowledge graphs for the learning task. As a result, their model can provide accurate social recommendations because it aggregates the social relationships of users and the interactions between users and items.

In addition, people's activities on social media reveal social relationships. For example, we can learn about the relationships around a person through his photos or comments on Twitter. Significantly, social relationship extraction assists companies in tracking users and enhancing the user experience. Therefore, many works are devoted to social relationship extraction. 
Wang et al. \citep{wang2018deep} propose a graph reasoning model to recognize the social relationships of people in a picture that is posted on social media. Their model enforces a particular function based on the social knowledge graph and deep neural networks. 
In their method, they initialized the relation edges and entity nodes with the features that are extracted from the semantic objects in an image. Then, they employed GGNN to propagate the knowledge graph. Therefore, they explored the relations of the people in the picture.


One of the biggest problems in this space is fake news \citep{zhang2019multi}. Online social media has become the principal platform for people to consume news. Therefore, a considerable amount of research has been done for fake news detection \citep{choi2020rumor,meel2020fake}. 
Most recently, Mayank et al. \citep{mayank2021deap} exploited a knowledge graph-based model called DEAP-FAKED to detect fake news on social media. Specifically, DEAP-FAKED learns news content and identifies existing entities in the news as the nodes of the knowledge graph. Afterward, a GNN-based technique is applied to encode the entities and detect anomalies that may be linked with fake news.

\subsection{Health/Medical Care}

With medical information explosively growing, medical knowledge analysis plays an instrumental role in different healthcare systems. Therefore, research focuses on integrating medical information into knowledge graphs to empower intelligent systems to understand and process medical knowledge quickly and correctly \citep{li2020real}. Recently, a variety of biomedical knowledge graphs have become available. Therefore, many medical care applications exploit knowledge graphs. For instance, Zhang et al. \citep{zhang2020hkgb} presented a Health Knowledge Graph Builder (HKGB) to build medical knowledge graphs with clinicians' expertise.

Specifically, we discuss the three most common intelligent medical care applications, including medical recommendation, health misinformation detection, and drug discovery. Firstly, with the rapid development of the medical industry, medical choices have become more abundant. Nevertheless, in the variety of medical choices, people often feel confused and unable to make the right decision to get the most suitable and personalized medical treatment. Therefore, medical recommender systems, especially biomedical knowledge graph-based recommender systems (such as doctor recommender systems and medicine recommender systems), have been put forward to deal with this issue \citep{katzman2018deepsurv}. Taking medicine recommendation as an example, Gong et al. \citep{gong2021smr} provided a medical knowledge graph embedding method by constructing a heterogeneous graph whose nodes are medicines, diseases, and patients to recommend accurate and safe medicine prescriptions for complicated patients.

Secondly, although many healthcare platforms aim to provide accurate medical information, health misinformation is an inevitable problem. Health misinformation is defined as incorrect information that contradicts authentic medical knowledge or biased information that covers only a part of the facts \citep{wang2020detecting}. Unfortunately, a great deal of health-related information on various healthcare platforms (e.g., medical information on social media) is health misinformation. What's worse, the wrong information leads to consequential medical malpractice; therefore, it is urgent to detect health misinformation. Utilizing authoritative medical knowledge graphs to detect and filter misinformation can help people make correct treatment decisions and suppress the spread of misinformation \citep{cui2020deterrent}. 
Representatively, Cui et al. \citep{cui2020deterrent} presented a model called DETERREN to detect health misinformation. DETERREN leverages a knowledge-guided attention network that incorporates an article-entity graph with a medical knowledge graph. 

Lastly, drug discovery, such as drug repurposing and drug-drug interaction prediction, has been a research trend for intelligent healthcare in recent years. Benefiting from the rich entity information (e.g., the ingredients of a drug) and relationship information (e.g., the interaction of drugs) in medical knowledge graphs, drug discovery based on knowledge graphs is one of the most reliable approaches \citep{maclean2021knowledge}.
Lin et al.  \citep{lin2020kgnn} presented an end-to-end framework called KGNN (Knowledge Graph Neural Network) for drug-drug interaction prediction. The main idea of KGNN is to mine the relations between drugs and their potential neighborhoods in medical knowledge graphs. It first exploits the topological information of each entity; then, it aggregates all the neighborhood information from the local receptive entities to extract both semantic relations and high-order structures. Wang et al. \citep{wang2020covid} developed a knowledge discovery framework called COVID-KG to generate COVID-19-related drug repurposing reports. They first constructed multimedia knowledge graphs by extracting medicine-related entities and their relations from images and texts. Afterward, they utilized the constructed knowledge graphs to generate drug repurposing reports.

\section{Technical Challenges}\label{sec5}

Although knowledge graphs offer fantastic opportunities for various services and applications, many challenges are yet to be addressed \citep{noy2019industry}. Specifically, the limitations of existing knowledge graph technologies are the key challenges for promoting the development of knowledge graphs \citep{hogan2021knowledge}. Therefore, this section discusses the challenges of knowledge graphs in terms of the limitations of five topical knowledge graph technologies, including knowledge graph embeddings, knowledge acquisition, knowledge graph completion, knowledge fusion, and knowledge reasoning.

\subsection{Knowledge Graph Embeddings}

\begin{sidewaystable}
	\sidewaystablefn%
	\begin{center}
		\begin{minipage}{\textheight}
			\caption{Knowledge graph embedding methods.}\label{embedding}
				\begin{tabular}{p{0.25\columnwidth}p{0.3\columnwidth}p{0.3\columnwidth}p{0.08\columnwidth}}
					\toprule
					Categories& Techniques & Evaluation Approaches\_Data Set& Results\\
					\midrule
					Tensor factorization-based methods&  RESCAL \citep{nickel2011three} & Link prediction[Hits@10]\_FB15K& 44.1\%\\
					 & HolE \citep{nickel2016holographic}& Link prediction[Hits@10]\_FB15K& 73.9\%\\
					 & ComplEx \citep{trouillon2016complex}& Link prediction[Hits@10]\_FB15K&84\%\\
					& SimplE \citep{kazemi2018simple}& Link prediction[Hits@10]\_FB15K& 83.8\%\\
					 & RotatE \citep{2019RotatE}& Link prediction[Hits@10]\_FB15K& 88.4\%\\
					 & QuatE \citep{2019Quaternion}& Link prediction[Hits@10]\_FB15K& 90\%\\
					Translation-based methods& TransE \citep{bordes2013translating}& Link prediction[Hits@10]\_FB15K& 47.1\%\\
					& TransH \citep{wang2014knowledge}& Link prediction[Hits@10]\_FB15K&64.4\%\\
					& TransR \citep{lin2015learning}&Link prediction[Hits@10]\_FB15K& 68.7\%\\
					& TransD \citep{ji2015knowledge}& Link prediction[Hits@10]\_FB15K& 77.3\%\\
					& TranSparse \citep{ji2016knowledge}& Link prediction[Hits@10]\_FB15K& 79.9\%\\
					& STransE \citep{nguyen2016stranse}& Link prediction[Hits@10]\_FB15K&79.7\%\\
					& TransA \citep{jia2016locally}& Link prediction[Hits@10]\_FB15K& 80.4\%\\
					& KG2E \citep{he2015learning}& Link prediction[Hits@10]\_FB15K& 71.5\%\\
					& TransG \citep{xiao2015transg}& Link prediction[Hits@10]\_FB15K& 88.2\%\\
					Neural network-based methods& SME \citep{bordes2014semantic} & Link prediction[Hits@10]\_FB15K&  41.3 \%\\
					& NTN \citep{socher2013reasoning} &Triplet classification[Accuracy]\_WN11&  86.2\%\\
					 & SLM \citep{socher2013reasoning}& Triplet classification[Accuracy]\_WN11&  76\%\\
					 & RMNN \citep{liu2016probabilistic} & Triplet classification[Accuracy]\_WN11&  89.9\%\\
					 & R-GCN \citep{schlichtkrull2018modeling}& Link prediction[Hits@10]\_FB15K& 84.2\%\\
					 & ConvKB \citep{nguyen2017novel}&Link prediction[Hits@10]\_WN18RR&  52.5 \%\\
					 & KBGAN \citep{cai2017kbgan}& Link prediction[Hits@10]\_WN18& 89.2\%\\
					\bottomrule
					
				\end{tabular}
		\end{minipage}
	\end{center}
\begin{tablenotes}
	\footnotesize
	\item{(1)} In this table, all the results of link prediction are filter results.

\end{tablenotes}
\end{sidewaystable}

The aim of knowledge graph embeddings is to effectively represent knowledge graphs in a low-dimensional vector space while still preserving the semantics \citep{Xia2021TAI,vashishth2020interacte}. 
Firstly, the entities and relations are embedded into a dense dimensional space in a given knowledge graph, and a scoring function is defined to measure the plausibility of each fact (triplet). Then, the plausibility of the facts is maximized to obtain the entity and relation embeddings \citep{chaudhri2022knowledge,sun2022relational}. The representation of knowledge graphs brings various benefits to downstream tasks. The three main types of triplet fact-based knowledge graph embedding approaches are tensor factorization-based, translation-based, and neural network-based methods \citep{rossi2021knowledge}.

\subsubsection{Tensor Factorization-based Methods}

\begin{figure}[!t]
	
	\centering
	\includegraphics[scale=0.5]{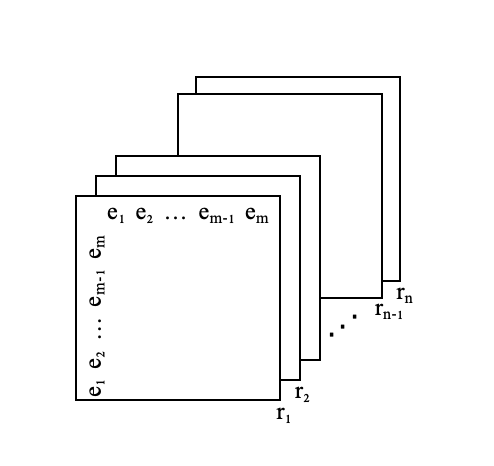}
	\caption{An illustration of tensor factorization of knowledge graphs.}
	\label{5}
\end{figure}

The core idea of tensor factorization-based methods is transforming the triplets in the knowledge graph into a $3$D tensor \citep{balavzevic2019tucker}. As Fig~\ref{5} presents, the tensor $\mathcal{X}\in R^{m\times m \times n}$, where $m$ and $n$ indicate the number of entity and relation, respectively, contains $n$ slices, and each slice corresponds to one relation type. If the condition $\mathcal{X}_{ijk}=1$ is met, the triplet $(e_i,r_k,e_j)$, where $e$ and $r$ denote entity and relation, respectively, exists in the knowledge graph. Otherwise, if $\mathcal{X}_{ijk}=0$, there is no such a triplet in the knowledge graph. Then, the tensor is represented by the embedding matrices that consist of the vectors of entities and relations.

\subsubsection{Translation-based Methods}
Translation-based methods exploit the scoring function, which is based on translation invariance. Translation invariance interprets the distance between the vectors of the two words, which is represented by the vector of their semantic relationships \citep{mikolov2013efficient}. Bordes et al. \citep{bordes2013translating} firstly utilized the translation invariance-based scoring functions to measure the embedding results. They creatively proposed the TransE model, which translates all the entities and relations of a knowledge graph into a continuous and low vector space. Specifically, the vectors of the head and tail entities in a triplet are connected by the vector of their relation. Consequently, in the vector space, the semantic meaning of every triplet is preserved. Formally, given a triplet $(head, relation, tail)$, the embedding vectors of the head entity, relation, and tail entity are $\mathbf{h}$, $\mathbf{r}$, and $\mathbf{t}$, respectively. In the vector space, the plausibility of the triplet $(\mathbf{h},\mathbf{r},\mathbf{t})$ is computed by the translation invariance-based scoring function to ensure it follows the geometric principle: $\mathbf{h}+\mathbf{r}\approx \mathbf{t}$.

After TransE, a lot of related extensions, such as TransH \citep{wang2014knowledge} and TransR \citep{lin2015learning}, are continually proposed to improve the performance of the Translation-based knowledge graph embeddings.

\subsubsection{Neural Network-based Methods}
Nowadays, deep learning has become a popular tool that is utilized for knowledge graph embeddings, and a considerable amount of research proposes to employ neural networks to represent the triplets of knowledge graphs \citep{dai2020generative}. 
In this section, we discuss three representative works, including SME, ConvKB, and R-GCN, to briefly introduce neural network-based knowledge graph embeddings.

SME \citep{bordes2014semantic} designs an energy function to conduct semantic matching, which utilizes neural networks to measure the confidence of each triplet $(h,r,t)$ in knowledge graphs. The scoring function of SME is defined as follows:
\begin{equation}
	f_r(h,t)=(\mathbf{W}_{h1}\mathbf{h}+\mathbf{W}_{h2}\mathbf{r}+\mathbf{b}_h)\top(\mathbf{W}_{t1}\mathbf{t}+\mathbf{W}_{t2}\mathbf{r}+\mathbf{b}_t).
\end{equation}
The scoring function of SME (bilinear) is:
\begin{equation}
	f_r(h,t)=((\mathbf{W}_{h1}\mathbf{h})\circ (\mathbf{W}_{h2}\mathbf{r})+\mathbf{b}_h)\top((\mathbf{W}_{t1}\mathbf{t})\circ (\mathbf{W}_{t2}\mathbf{r})+\mathbf{b}_t).
\end{equation}
Here $\mathbf{W} \in \mathbb{R}^{d\times d}$ denotes the weight matrix, $\mathbf{b}$ indicates the bias vector. $\mathbf{h}$, $\mathbf{r}$, and $\mathbf{t}$ are the embedding vectors of head entity, relation, and tail entity, respectively.

ConvKB \citep{nguyen2017novel} utilizes a convolutional neural network (CNN) to conduct knowledge graph embeddings. ConvKB represents each triplet $(h, r, t)$ as a three-row matrix $\mathbf{A}$, which is input to a convolution layer to obtain feature maps. Afterward, the feature maps are concatenated as a vector, and then a score is calculated to estimate the confidence of the triplet. The scoring function is as follows:
\begin{equation}
	f_r(h,t)=O(\textsl{g}(\mathbf{A}*\Omega))\mathbf{w},
\end{equation}
where $O$ signifies the concatenation operator, $\textsl{g}(\cdot)$ is the ReLU activation function, $\mathbf{A}*\Omega$ indicates the convolution operation of matrix $\mathbf{A}$ by using the filters in the set $\Omega$, $\mathbf{w}\in \mathbb{R}^{3d}$ is a weight vector.

R-GCN \citep{schlichtkrull2018modeling} is an improvement of graph neural networks (GNNs). R-GCN represents knowledge graphs by providing relation-specific transformation. Its forward propagation is calculated as follows:
\begin{equation}
	h_k^{(l+1)}=\sigma\bigg(\sum_{r\in R}\sum_{i\in N_k^r}\frac{1}{n_{k,r}}W_i^{(l)}h_i^{(l)} +W_k^{(l)}h_k^{(l)} \bigg),
\end{equation}
where $h_k^{(l+1)}$ is the hidden state of the entity $k$ in $l$-th layer, $N_k^r$ denotes
a neighbor collection of entity $k$ and relation $r \in R$, $n_{k,r}$ is the normalization process, $W_i^{(l)}$ and $W_k^{(l)}$ are the weight matrices.

\subsubsection{Limitations of Existing Methods}

The existing methods for generating knowledge graph embeddings still suffer several severe limitations. Many established methods only consider surface facts (triplets) of knowledge graphs. However, additional information, such as entity types and relation paths, are ignored, which can further improve the embedding accuracy. The performance of most traditional methods that do not consider the additional information is unsatisfactory.
Table~\ref{embedding} lists the embedding methods, which do not consider the additional information. In Table~\ref{embedding}, the performance evaluation is based on the link prediction and triplet classification tasks. The metrics that are for evaluation results are hit rate at 10 (Hits@10) and accuracy. 
As Table~\ref{embedding} presents, only a few models have impressive results, including the results of QuatE (90\%), RMNN (89.9\%), and KBGAN (89.2\%). 
Recently, some researchers have started to combine additional information with a knowledge graph to improve the efficiency of embedding models. For example, Guo et al. \citep{guo2015semantically} take advantage of additional entity type information, which is the semantic category of each entity, to obtain the correlation between the entities and to tackle the data sparsity issue. Therefore, knowledge graphs are represented more accurately. Not only entity types, some other information, including relation paths \citep{li2021learning}, time information of dynamic graphs \citep{messner2022temporal}, and textual descriptions of entities \citep{an2018accurate}, are getting the researchers' attention in recent years. However, it is still a daunting challenge to effectively utilize rich additional information to improve the accuracy of knowledge graph embeddings.

General additional information can not adequately represent the semantic meaning of the triplets. For instance, the entity types are not related to the semantic information of triplets.
Furthermore, the types of additional information that can be incorporated into the features of the triplets are now severely limited. Therefore, to improve the performance of existing knowledge graph embedding methods, multivariate information (such as the hierarchical descriptions of relations and the combination of entity types and textual descriptions) needs to be incorporated into the features of the triplets.

 To the best of our knowledge, complex relation path remains an open research problem \citep{peng2021knowledge}. For example, the inherent relations, referring to the indirect relationships between two unconnected entities, are not represented effectively. Although the inherent relations between the entities can be explored based on the chain of relationships in knowledge graphs, the inherent relations are complex and multiple. Therefore, it is not straightforward to represent these relations effectively.

\subsection{Knowledge Acquisition}

Knowledge acquisition is a critical step for combining data from different sources and generating new knowledge graphs. The knowledge is extracted from both structured and unstructured data. 
Three main methods of knowledge acquisition are relation extraction, entity extraction, and attribute extraction \citep{fu2019graphrel}. Here, attribute extraction can be regarded as a special case of entity extraction. Zhang et al. \citep{zhang2019long} took advantage of knowledge graph embeddings and graph convolution networks to extract long-tail relations. Shi et al. \citep{shi2021entity} proposed entity set expansion to construct large-scale knowledge graphs.

Nevertheless, existing methods for knowledge acquisition still face the challenge of low accuracy, which could result in incomplete or noisy knowledge graphs and hinder the downstream tasks. Therefore, the first critical issue regards the reliability of knowledge acquisition tools and their evaluation. In addition, a domain-specific knowledge graph schema is knowledge-oriented, while a constructed knowledge graph schema is data-oriented for covering all data features \citep{zhou2022schere}. Therefore, it is inefficient to produce domain-specific knowledge graphs by extracting entities and properties from raw data. Hence, it is an essential issue to efficiently achieve knowledge acquisition tasks by generating domain-specific knowledge graphs.

Besides, most existing knowledge acquisition methods focus on constructing knowledge graphs with one specific language. However, in order to make the information in knowledge graphs richer and more comprehensive, we need cross-lingual entity extraction. It is thus vitally important to give more attention to cross-lingual entity extraction and the generation of multilingual knowledge graphs. For example, Bekoulis et al.\citep{bekoulis2018joint} proposed a joint neural model for cross-lingual (English and Dutch) entity and relation extraction. Nevertheless, multilingual knowledge graph construction is still a daunting task since non-English training data sets are limited, language translation systems are not always accurate, and the cross-lingual entity extraction models have to be retrained for each new language.

Multi-modal knowledge graph construction is regarded as another challenging issue of knowledge acquisition. The existing knowledge graphs are mostly represented by pure symbols, which could result in the poor capability of machines to understand our real world \citep{zhu2022multi}. Therefore, many researchers focus on multi-modal knowledge graphs with  various entities, such as texts and images. The construction of multi-modal knowledge graphs requires the exploration of entities with different modalities, which makes the knowledge acquisition tasks complicated and inefficient.

\subsection{Knowledge Graph Completion}

Knowledge graphs are often incomplete, i.e., missing several relevant triplets and entities \citep{zhang2020learning}. For instance, in Freebase, one of the most well-known knowledge graphs, more than half of person entities do not have information about their birthplaces and parents. 
Generally, semi-automated and human leveraging mechanisms, which can be applied to ensure the quality of knowledge graphs, are essential tools for the evaluation of knowledge graph completion. Specifically, human supervision is currently considered the gold standard evaluation in knowledge graph completion \citep{ballandies2021mobile}.

Knowledge graph completion aims to expand existing knowledge graphs by adding new triplets using techniques for link prediction \citep{Wang2020TCSSmodel,akrami2020realistic} and entity prediction \citep{ji2021survey}. 
These approaches typically train a machine learning model on the knowledge graph to assess the plausibility of new candidate triplets. Then, they add the candidate triplets with high plausibility to the graph. 
For example, for an incomplete triplet \emph{(Tom, friendOf, ?)}, it is possible to assess the range of tails and return the more plausible ones to enrich the knowledge graph. These models successfully utilized knowledge graphs in many different domains, including digital libraries~\citep{yao2017incorporating}, biomedical~\citep{harnoune2021bert}, social media~\citep{abu2021domain}, and scientific research~\citep{nayyeri2021trans4e}. 
Some new methods are able to process fuzzy knowledge graphs in which each triple is associated with a confidence value \citep{chen2019embedding}. 

However, most current knowledge graph completion methods only focus on extracting triplets from a closed-world data source. That means the generated triplets are new, but the entities or relations in the triplets need to already exist in the knowledge graph. 
For example, for the incomplete triplet \emph{(Tom, friendOf, ?)}, predicting the triplet \emph{(Tom, friendOf, Jerry)} is only possible if the entity \emph{Jerry} is already in the knowledge graph.
Because of this limitation, these methods cannot add new entities and relations to the knowledge graph. To tackle this issue, we are starting to see the emergence of open-world techniques for knowledge graph completion that extracts potential objects from outside of the existing knowledge bases. For instance, the ConMask model \citep{shi2018open} has been proposed to predict the unseen entities in knowledge graphs. However, methods for open-world knowledge graph completion still suffer from low accuracy. The main reason is that the data source is usually more complex and noisy. In addition, the similarity of the predicted new entities to the existing entities can mislead the results. In other words, two similar entities are regarded as connected entities, while they may not have a direct relationship.

Knowledge graph completion methods assume knowledge graphs are static and fail to capture the dynamic evolution of knowledge graphs. To obtain accurate facts over time, temporal knowledge graph completion, which considers the temporal information reflecting the validity of knowledge, has emerged. Compared to static knowledge graph completion, temporal knowledge graph completion methods integrate timestamps into the learning process. Hence, they explore the time-sensitive facts and improve the link prediction accuracy significantly. Although temporal knowledge graph completion methods have shown brilliant performance, they still face serious challenges. Because these models consider time information would be less efficient \citep{shao2022tucker}, the key challenge of temporal knowledge graph completion is how to effectively incorporate timestamps of facts into the learning models and properly capture the temporal dynamics of facts.

\subsection{Knowledge Fusion}

Knowledge fusion aims to combine and integrate knowledge from different data sources. It is often a necessary step for the generation of knowledge graphs \citep{nguyen2020knowledge,smirnov2019knowledge}. The primary method of knowledge fusion is entity alignment or ontology alignment \citep{Ren2021TETCI}, which aims to match the same entity from multiple knowledge graphs \citep{zhao2020multi}. 
Achieving efficient and accurate knowledge graph fusion is a challenging task because of the complexity, variety, and large volume of data available today.

While a lot of work has been done in this direction, there are still several intriguing research directions that deserve to be investigated in the future. One of them regards cross-language knowledge fusion \citep{mao2020mraea}, which allows the integration of information from different languages. This is often used to support cross-lingual recommender systems \citep{javed2021review}. 
For example, Xu et al. \citep{xu2019cross} adopted a graph-matching neural network to achieve cross-language entity alignment. However, the result of the cross-language knowledge fusion is still unsatisfactory because the accuracy of the matching entities from different languages is relatively low. Therefore, it remains a daunting challenge to explore cross-language knowledge fusion.

Another primary challenge regards entity disambiguation  \citep{nguyen2020knowledge}. As the polysemy problem of natural language, the same entity may have various expressions in different knowledge graphs. Hence, entity disambiguation is required before conducting entity alignment. Existing entity disambiguation methods mainly focus on discriminating and matching ambiguous entities based on extracting knowledge from texts containing rich contextual information \citep{zhu2018exploiting}. However, these methods can not precisely measure the semantic similarity of entities when the texts are short and have limited contextual information. Only a few works have focused on solving this issue. For example, Zhu and Iglesias \citep{zhu2018exploiting} have proposed SCSNED for entity disambiguation. 
SCSNED measures semantic similarity based on both informative words of entities in knowledge graphs and contextual information in short texts. Although SCSNED alleviates the issue of limited contextual information to some extent, more effort is needed to improve the performance of entity disambiguation.

In addition, many knowledge fusion methods only focus on matching entities with the same modality and ignore multi-modal scenes in which knowledge is presented in different forms. Specifically, entity alignment considering only single-modality knowledge graph scenario has insignificant performance because it can not fully reflect the relationships of entities in the real world \citep{cheng2022multijaf}. Recently, to solve this issue, some studies have proposed multi-modal knowledge fusion, which matches the same entities having different modalities and generates a multi-modal knowledge graph. For example, HMEA \citep{guo2021multi} aligns entities with multiple forms by mapping multi-modal representations into hyperbolic space. Although many researchers have worked on multi-modal knowledge fusion, it is still a critical task. Multi-modal knowledge fusion mainly aims to find equivalent entities by integrating their multi-modal features \citep{cheng2022multijaf}. Nevertheless, how to efficiently incorporate the features having multiple modalities is still a tricky issue facing current methods.

\subsection{Knowledge Reasoning}

The goal of knowledge reasoning is to infer new knowledge, such as the implicit relations between two entities \citep{liu2019TKDE,wang2019explainable}, based on existing data. For a given knowledge graph, wherein there are two unconnected entities $h$ and $t$, denoted as $h,t\in G$, here $G$ means the knowledge graph, knowledge reasoning can find out the potential relation $r$ between these entities and form a new triplet $(h,r,t)$. 
The knowledge reasoning methods are mainly categorized into logic rule-based \citep{de2021rdf}, distributed representation-based \citep{chen2020review}, and neural network-based methods \citep{xiong2017deeppath}. Logic rule-based knowledge reasoning aims to discover knowledge according to the random walk and logic rules, while distributed representation-based knowledge reasoning embeds entities and relations into a vector space to obtain distributed representation \citep{chen2020review}. Neural network-based knowledge reasoning method utilizes neural networks to infer new triplets given the body of knowledge in the graph \citep{xian2019reinforcement}. 

There are two tasks in knowledge reasoning: single-hop prediction and multi-hop reasoning \citep{ren2022smore}. Single-hop prediction predicts one element of a triplet for the given two elements, while multi-hop reasoning predicts one or more elements in a multi-hop logical query. In other words, in the multi-hop reasoning scenario, finding the answer to a typical question and forming new triplets requires the prediction and imputation of multiple edges and nodes. Multi-hop reasoning achieves a more precise formation of triplets when compared with the single-hop prediction. Therefore, multi-hop reasoning has attracted more attention and become a critical need for the development of knowledge graphs in recent years. Although many works have been done, multi-hop reasoning over knowledge graphs remains largely unexplored. Notably, multi-hop reasoning on massive knowledge graphs is one of the challenging tasks \citep{zhu2022step}. For instance, most recent studies focus on multi-hop reasoning over knowledge graphs, which have only 63K entities and 592K relations. The existing models can't learn the training set effectively for a massive knowledge graph that has more than millions of entities. Moreover, multi-hop reasoning needs to traverse multiple relations and intermediate entities in the knowledge graph, which could lead to exponential computation cost \citep{zhang2021cone}. Therefore, it is still a daunting task to explore multi-hop knowledge reasoning.

Besides, the verification of inferred new knowledge is also a critical issue. Knowledge reasoning enriches existing knowledge graphs and brings benefits to the downstream tasks \citep{wan2021reasoning}. However, the inferred new knowledge is sometimes uncertain, and the veracity of new triplets needs to be verified. Furthermore, the conflicts between new and existing knowledge should be detected. To address these problems, some research has proposed multi-source knowledge reasoning \citep{zhao2020multi} that detects erroneous knowledge and conflicting knowledge. Overall, more attention should be paid to multi-source knowledge reasoning and erroneous knowledge reduction.

\section{Conclusion}\label{sec6}

Knowledge graphs have played an instrumental role in creating many intelligent services and applications for various fields. In this survey, we provided an overview of knowledge graphs in terms of opportunities and challenges. We first introduced the definitions and existing research directions regarding knowledge graphs to provide an introductory analysis of knowledge graphs. Afterward, we discussed AI systems that take advantage of knowledge graphs. Then, we presented some representative knowledge graph applications in several fields. 
Furthermore, we analyzed the limitations of current knowledge graph technologies, which lead to severe technical challenges. We expect this survey to spark new ideas and insightful perspectives for future research and development activities involving knowledge graphs.


\section*{Declarations}

\bmhead{Conflict of interest} The authors declare that they have no competing financial interests or personal relationships that could have appeared to influence the work reported in this paper.

\bibliographystyle{sn-basic.bst}
\bibliography{Bibfile}


\end{document}